# On Developing Facial Stress Analysis and Expression Recognition Platform

Fabio Cacciatori, *Illogic S.r.l, cacciatori@illogic.xyz*,
Sergei Nikolaev, *Cyberphysics LLC, s.nikolaev@cyberphysics.xyz*,
Dmitrii Grigorev, *Illogic S.r.l., dm.srg.gr@gmail.com*,
Anastasiia Archangelskaya, *Skolkovo Institute of Science and Technology, anastasiia.archangelskaya@skoltech.ru*

*Abstract*—This work represents the experimental and development process of system facial expression recognition and facial stress analysis algorithms for an immersive digital learning platform. The system retrieves from users web camera and evaluates it using artificial neural network (ANN) algorithms. The ANN output signals can be used to score and improve the learning process. Adapting an ANN to a new system can require a significant implementation effort or the need to repeat the ANN training. There are also limitations related to the minimum hardware required to run an ANN. To overpass these constraints, some possible implementations of facial expression recognition and facial stress analysis algorithms in real-time systems are presented. The implementation of the new solution has made it possible to improve the accuracy in the recognition of facial expressions and also to increase their response speed. Experimental results showed that using the developed algorithms allow to detect the heart rate with better rate in comparison with social equipment.

## I. INTRODUCTION

Avatar.xyz [1] is a software for the creation of "Serious Games"; informally defined as "video games with a purpose that is not exclusively playful", for example for educational purposes. Avatar allows users to immerse in a virtual training experience. Thanks to the use of innovative technologies such as artificial intelligence, computer vision algorithms, and virtual assets (people digital twins or avatar, and scenographies), the users are able to experience a situational or face-to-face training simulation (i.e. synthetic training), according to their needs or indications from their employers. AVATAR is composed of a Back-Office platform designed for the management, creation, and modification of training contents, and the Player, which allows the user to carry out the courses.

Stress negatively affects the learning process. When a student receives a new task, he becomes anxious. If the task is too complex and large, fatigue sets in. The brain can no longer resist stress, and the student ceases to be productive. For this reason, the process of monitoring student stress and the emotional state plays a huge role in the learning process.

One of the main objectives of the application is to bring innovative training experience to users and make the training process more engaging and therefore more efficient in terms of storing shared contents and learning. The Avatar.xyz software uses artificial intelligence algorithms based on neural networks to implement four features: speech recognition, facial expression classification, stress level analysis, and attention level measurement. This paper considers two of these features: facial stress analysis (FSA) and facial expression recognition (FER) algorithms.

The FSA algorithm is able to analyze - applying a specific color filter - the flush of blood after every heartbeat and estimate the number of heartbeats of the user, thus also allowing an estimate of the level of stress of the learner who is taking the course.

The facial expression recognition algorithm implemented in the AVATAR platform allows the recognition of up to 7 different facial expressions and, if the trainer wishes, to influence the learner's training path based on these results
Heart rate (HR).

There exists a problem of emotion and stress recognition in order to enhance the whole educational process and define effective parts of the course. The approach developed in [2] combines spatial and temporal processing to emphasize subtle temporal changes in a video. In [6] the algorithm with basic facial emotions recognition was described, where the CNN network takes a 48x48 face as input and returns a label between 0 and 6 according to the scheme of the FER-2013 dataset. The author reports an accuracy in identifying the expressed emotion of approximately 66% using the FER-2013 dataset.

In this paper we proposed facial emotions recognition (FER) and facial stress analysis (FSA) algorithms integrated into the Unity and course platform built on Angular framework in order track user reactions during the course completion. Our FER algorithm showed an accuracy of 70.1%, about 4% higher than the described approaches.

## II. SYSTEM ARCHITECTURE

For the development of the Avatar platform, a Microservices architecture has been adopted. Each component was developed independently and the http communication protocol is used for the exchange of information between the various services.

The Avatar API manages the logic of the platform (Fig. 6). In addition, this component manages communication between the various subsystems.

The API handles the business logic and the persistence of data like media files and course data. As shown in the figure below, communication between client and server is done using the http protocol and the json format for message exchange.

### III. FSA Algorithm Description

Stress level analysis is derived directly from an estimate of the participant's heart rate. To make the system easier and cheaper to deploy, the heart rate measurement is not done through wearable devices, but rather through an analysis of facial skin coloration. A computer vision algorithm can notice the sudden change in the coloration of the face (invisible to the human eye) due to the influx of blood as a result of the heartbeat. To date, no avatar-based training solutions are using this technique.

Heart rate (HR) is an important indicator of people's physiological state. Recent research has shown that these physiological signals can be measured with a simple camera. The human visual system has limited spatial-temporal sensitivity, but many signals that do not fall within this limitation can be informative. For example, human skin color varies slightly with blood circulation. This variation, although invisible to the naked eye, can be used to extract a person's heart rate.

The idea of monitoring physiological parameters without sensors worn by the person is to be found in the cardiovascular system of the human body. The cardiovascular system allows blood to circulate in the body thanks to the continuous pumping of blood from the heart. Our heart pumps blood through the blood vessels, and with each heartbeat, the circulation of the blood creates a color change in the skin of the face. Therefore, it is possible to extract HR from the change in the skin color of the face. We are nowadays experimenting as well with the possibility to calculate blood pressure.

The approach developed in [2] combines spatial and temporal processing to emphasize subtle temporal changes in a video. First, the video sequence is decomposed into different spatial frequency bands. These bands may be magnified differently because they may have different signal-to-noise ratios, or they may contain spatial frequencies for which the linear approximation used in the motion magnification may not be suitable. In the latter case, the amplification for these bands is reduced to suppress artifacts. The goal of spatial processing is, therefore, simply to increase the temporal signal-to-noise ratio by putting together more pixels, spatially filtering the frames, and applying downsampling for computational efficiency. In the specific case, however, a complete Laplacian pyramid is calculated as in [3].

The second phase involves performing temporal processing on each space band. The time-series corresponding to the value of a pixel in a frequency band, are considered and a band-pass filter is applied to extract the frequency of the band of interest. For example, frequencies within 0.4-4Hz could be selected, which is approximately 24-240 beats per minute.

The temporal processing is uniform for all spatial layers and all pixels within each layer. The extracted signal is then multiplied by a magnification factor called α. This factor can be specified by the user or can be obtained automatically.

In the next phase, the zoomed signal is added to the original signal and collapsed space pyramid to get the final output. For further details about the mathematics behind this system, please refer directly to the paper [2].

The technique can be performed in real-time to show phenomena occurring at user-selected time frequencies. The developed version has the following changes:

1. Heart rate monitoring via webcam.
2. Face detection algorithm implementation.
3. Tuning of some computation parameters.

As for the first point, the beat calculations were inserted directly from the webcam. In this way, if the debug mode is activated, a window is displayed in which the image captured by the camera is represented and the computed BPM value is reported. The algorithm needs an initial phase equivalent to about 5 seconds for the initialization of the calculations and to keep the output BPM value more stable.

Face detection is performed using the algorithm in [4]. This method is very fast and quite accurate in the calculation. It could also be replaced with a CNN network: this option is under evaluation since this system has been integrated into the Avatar platform. Face detection is as well used to identify if there is a face within the area to be inspected. Intersection over union metric is used to decide if a face is present within the area in question. This metric works very well, and we have decided to implement it according to this scheme: our ground truth is represented by the inspection area for the recognition of the beat while the recognized face is our identified object. At this point, if the intersection between the union has a ratio greater than a certain threshold, the beat is computed, otherwise, that frame is not considered and a new face is expected to be identified.

The screen, in debug mode, is shown in figure 1. In green is the area being analyzed, and in blue is the result of the face detector.

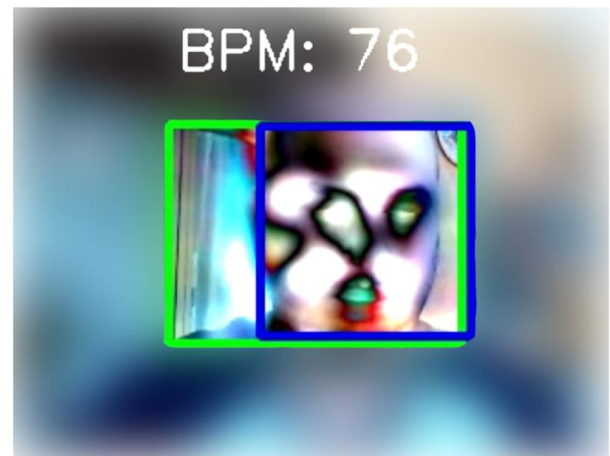

**Fig. 1.** Debug mode with display of the output window of the developed code.

## IV. FER Algorithm Description

Facial Expression Recognition is a feature that allows you to monitor the user's facial expression. It is based on Computer Vision algorithms able to identify seven different facial expressions (neutral, angry, disgust, fear, happy, sad, and surprised) and indicate the level of certainty.

The facial expression recognition system allows to perform an analysis of the emotional state of a user who performs the course during their training session. In particular, the system must capture the webcam stream and recognize the expression of the user's face. The seven emotional states are:

- Angry
- Disgust
- Fear
- Happy
- Sad
- Surprise
- Neutrale

The software module that manages the FER was written using the Python language. In addition, the "Keras" and OpenCV Machine Learning libraries were used.
The algorithm that detects the user's expression consists of three main components:

- REST interface to receive the input frame to analyze and to return the output;
- Python class that uses OpenCV's Viola-Jones algorithm to identify the user's face;
- Neural network implemented in Keras that takes previously identified face and classifies its expression.

The reference dataset is called FER-2013 [5]. The data consists of 48x48 pixel grayscale face images. The faces have been registered automatically so that the face is more or less centered and takes up about the same amount of space in each image.

The training dataset consists of 28,709 examples. The public test dataset used for ranking consists of 3,589 examples. The final test dataset, which was used to determine the contest winner, consists of another 3,589 examples.

In [6], Global Average Pooling is used to eliminate the fully connected layer. This was possible by having in the last convolutional layer the same number of feature maps as the number of classes and by applying a SoftMax activation function to each reduced feature map. In this way, their initial architecture is, in fact, a fully convolutional standard neural network consisting of 9 layers of convolution, ReLU, batch normalization, and Global Average Pooling and contains approximately 600,000 parameters.

Since our initially proposed architecture completely eliminated the last connected layer, we have further reduced the number of parameters by now eliminating them from the convolutional layers.

This was done using depth-wise separable convolutions [7]. Depth-wise separable convolutions are composed of two different layers: depth-wise convolutions and point-wise convolutions. The main purpose of these layers is to separate the spatial cross-correlations from the channel cross-correlations [6]. It was achieved by applying a DxD filter on each M input channel and then applying N convolution size 1x1xM filters to combine the M input channels into N output channels. The application of 1x1xM convolutions combines each value in the feature map without considering their relationship space within the channel.

The final neural network architecture contains four residual depth-wise separable convolutions where each convolution is followed by batch normalization and a ReLU activation function. The output layer applies a global average pooling and a soft-max activation function to make a prediction. This architecture has about 60,000 parameters, which corresponds to a 10-fold reduction compared to the originally proposed architecture. The final weights of the architecture are stored in an 855 Kilobytes file. Furthermore, everything happens in real-time with a time of about 0.22 seconds (also considering face recognition).

The following is a description of several architectural improvements to the current system that allows better results in emotion detection.

The first option is to use one of the following solutions [8, 9]. These solutions implement VGG network architectures [10]. These two solutions achieved a result of 70.382% and 73.112% accuracy on the FER-2013 dataset respectively. Unfortunately, the calculation times of this network are unknown as they are not reported and, therefore, it is a question of testing the network to verify the actual real-time of the system.

Another way involves creating a new network starting, perhaps, from one of those previously seen and then verifying the results that can be obtained by comparing them with other state-of-the-art networks. This solution was chosen for the final implementation. This solution had the most advantages from the point of view of practicality and implementation speed. In addition, it allows one to have full control of the network and not rely on a network developed by a third party. After several attempts to try to optimize the network model to be created, we then implemented a CNN network which is then shown in figure three.

The confusion matrix obtained after training presented in Table 1.

TABLE I
CONFUSION MATRIX

|  | Angry | Disgust | Fear | Happy | Sad | Surprise | Neutral |
|---|---|---|---|---|---|---|---|
| Angry | 0.63 | 0.01 | 0.06 | 0.02 | 0.16 | 0.01 | 0.10 |
| Disgust | 0.20 | 0.69 | 0.02 | 0.00 | 0.04 | 0.04 | 0.02 |
| Fear | 0.12 | 0.00 | 0.43 | 0.02 | 0.18 | 0.10 | 0.13 |
| Happy | 0.02 | 0.00 | 0.01 | 0.90 | 0.02 | 0.02 | 0.03 |
| Sad | 0.07 | 0.00 | 0.08 | 0.04 | 0.58 | 0.02 | 0.21 |
| Surprise | 0.02 | 0.00 | 0.07 | 0.06 | 0.02 | 0.81 | 0.03 |
| Neutral | 0.04 | 0.00 | 0.03 | 0.04 | 0.13 | 0.01 | 0.75 |

The first integration was done by encapsulating a model proposed by [6] that managed to achieve 66% accuracy with two models. The first model uses about 600,000 parameters, while the second one, inspired by Xception [7], uses only 60,000 parameters. This second model is therefore preferable for performance reasons and succeeds in performing the classification in only 0.22 seconds on a medium-performance PC.

A repeated cycle of configuration and training performed on VGG16 and VGG1928 architectures allowed accuracy of 70.1%.

It is important to note that an accuracy value of 70%, although far from the desired 100%, is actually a very good result. In fact, in a task of discrimination among 7 different membership classes, a purely random selection would result in a starting accuracy of "one-seventh", i.e., about 14.3%.

## V. FSA ALGORITHM TESTS

The FSA algorithm tests were divided into 2 phases. These tests have been conducted on the system to verify its correctness and therefore analyze the results obtained.

In the first phase, three three-minute videos were recorded in 3 different phases of the morning (to have slightly different lighting conditions). The first 30 seconds of the videos were used to initialize the two systems. The results were analyzed by comparing the developed algorithm data with the data extracted from an Apple Watch Series 3. The heart rate data was extracted every 10 seconds during video recording and subsequently reported in tables to be able to compare the graphs.

In the second phase, three three-minute videos were recorded during the afternoon (also in this case to test different lighting conditions). Again, the first 30 seconds of the videos were used to initialize the two systems. The results were analyzed by comparing the data of the developed algorithm with the data extracted from an oximeter that allows for evaluating the oxygen saturation of the hemoglobin present in the peripheral arterial blood (defined with the abbreviation "SpO2") and, at the same time, allows to also measure a person's heart rate. This tool is even more accurate than the Apple Watch used in previous tests and, for this reason, we believe it can also provide more accurate data on the goodness of the algorithm itself.

Some functional tests of possible integration of the FSA algorithm inside a moveable camera were also included. Compared to the tests seen in phases 1 and 2, these provide for a distance of the face from the camera equal to 1.5 / 2 meters. For this reason, we carried out 2 tests (one at a distance of 1.5 m and one at a distance of 2 m) to evaluate the reliability of the code and discussed the results.

In Table 2 we report the comparison values between the averages calculated on the three tests between the two approaches (Apple Watch and developed algorithm). In particular, the last column shows the average error difference.

TABLE II
COMPARISON BETWEEN THE AVERAGES OF THE TWO APPROACHES WITH THE MEAN DEVIATION OF ERROR

| #Test | Method | Average bpm | Offset bpm |
|---|---|---|---|
| 1 | AW | 84.8125 | 5.4375 |
| 1 | AL | 79.375 | 5.4375 |
| 2 | AW | 86.5 | 3.8125 |
| 2 | AL | 82.6875 | 3.8125 |
| 3 | AW | 80.8125 | 0.3125 |
| 3 | AL | 80.5 | 0.3125 |
| Total Average | | 3.1875 | |

The data in Table 2 show an error between the two algorithms of about 3.1 beats per minute. In general, the data estimated by the algorithm are always lower than those measured by the Apple Watch. However, the comparison is encouraging. The average absolute error between the two approaches is 3.1875 beats. For the system that is to be achieved, this gap turns out to be a good value to be able to carry out the analysis.

The difference can also be explained as a function of the different brightness, the more or less still positions, and the head movement of the subject on which the measurement is performed. In any case, this also helps us to better understand how the algorithm behaves in different conditions.

During the tests, moments were verified in which the algorithm greatly underestimates the value of the beat (e.g. 30-40 bpm) this could be due to some error in the calculation of the beats or to a problem related to the area in which the computation is carried out. Fortunately, the values return to normal after a few seconds. We would like to underline that this behavior occurs very rarely and that, for this reason, the tests were conducted by measuring the heartbeat every 10 seconds.

As in the previous example, in Table 2 we report the comparison values between the averages calculated on the three tests between the two approaches (Pulse oximeter and algorithm developed). In particular, the last column shows the average error difference.

TABLE III
COMPARISON BETWEEN THE AVERAGES OF THE TWO APPROACHES WITH THE MEAN ERROR DIFFERENCE

| #Test | Method | Average bpm | Offset bpm |
|---|---|---|---|
| 1 | S | 81.125 | 5.4375 |
| 1 | AL | 79.4375 | 5.4375 |
| 2 | S | 78.25 | 3.8125 |
| 2 | AL | 78.875 | 3.8125 |
| 3 | S | 79.3125 | 0.3125 |
| 3 | AL | 78.875 | 0.3125 |

| | |
|---|---|
| **Total Average** | 3.1875 |

The data in table 2 show an error between the two algorithms of about 1 beat. The error, in this case, is lower than the one calculated via Apple Watch, probably also due to a different evaluation between the apple watch and oximeter or to the different lighting conditions of the subject since the test was conducted in the afternoon.

In general, the data are even more encouraging than the previous ones and this is an excellent signal on the goodness of the algorithm developed. The oscillations of the oximeter between the minimum and maximum calculated frequency are lower than those of the Apple Watch and this represents a better approximation of the calculation by the oximeter itself. Even the developed algorithm, as highlighted in the previous tests, sometimes has peaks but the values are uniform by calculating the average over longer times.

As described in the introduction, this paragraph will analyze the results obtained with our algorithm having a camera-face distance of 1.5 and 2 meters. For this reason, we have carried out two tests to be able to verify the efficiency of this algorithm in these 2 limit situations. The videos are always 3 minutes long and are analyzed starting from minute 2:30 to stabilize the algorithm calculations. The oximeter was used as a comparison approach as it is more accurate than the Apple Watch. The videos were recorded in the morning with sunlight.

To carry out these new tests, we had to modify some parameters of existing solutions. First, we had to eliminate the face detector. This is because, at distances greater than 1.2 meters, the face detection algorithm is no longer able to find faces (the faces are too small given the resolution of 320x240). The alternative would be to increase the size of the image, but this can lead to a slowdown in the execution of the algorithm itself. Several parameters of the Haar Cascade (face detection algorithm) have been tested but we have not been able to overcome this problem. We also tried using a CNN network (MTCNN) but this too has the same limitations. From an estimate made, the ratio of the face area to the heart rate calculation area is about 10%. As this ratio is very small, we believe that much of the heart rate calculation is related to misreading background information. For this reason, we have added a heart rate measurement carried out on background only, therefore without a human face, to understand if this hypothesis can be confirmed.

While the trend of the oximeter is quite constant, with a difference between the minimum and maximum, for the two experiments, equal to 10 and 13 beats respectively (1.5 meters and 2 meters), the algorithm data show oscillations similar to those of the background only, with differences between minimum and maximum equal to 32 and 38 (1.5 meters and 2 meters), equal to three times the real value read by the oximeter. These values turn out to be much higher than in the previous tests and, combined with the fact that the beats trend is similar to the one read from the background, making us assume that the measurement is strongly conditioned by the latter and that, therefore, the value that we read is not that of the person's heartbeat.

Even by increasing the resolution, this algorithm cannot be used at such distances because the calculation area is always greater than the face area and therefore the background will increasingly influence the heart rate value.

Alternatively, one could think of calculating the heartbeat only in the face area but due to the structure of the algorithm itself (the Gaussian pyramid), the face should always be resized to a standard size before making the calculations. Furthermore, we would still be bound by the maximum face recognition distance of the face detector.

A series of tests were also carried out on videos extracted directly from the moveable camera. The data processed by the algorithm were compared with those extrapolated from the apple watch. The beat values were calculated considering all the 30 seconds of acquisition and not every 10 as occurred in the previous experiments. The results confirmed what was analyzed with the last 2 tests carried out: if the person is at a distance greater than 1.2 meters, the face detector struggles to find the face (since it turns out to be very small) and the pulse value is strongly influenced by the background. Some screenshots are shown in the following figures.

In Figure 2, the subject is about 2 meters away and an instant beat of 90 bpm is identified when the real value is about 60. The average on the video is 75 versus the real 63 bpm. As you can see, almost the entire torso of the subject falls in the green analysis box and this, unfortunately, creates several problems for the algorithm calculation.

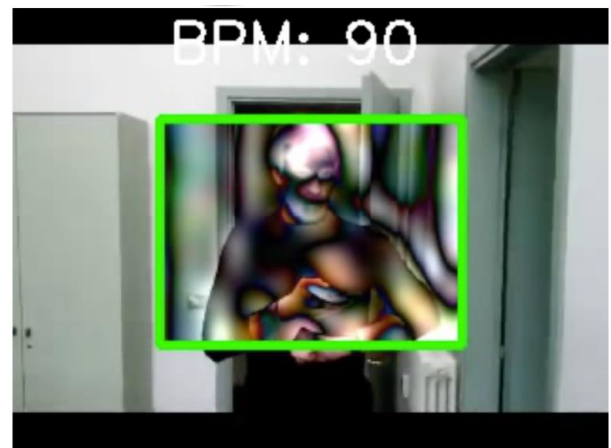

**Fig. 2.** Moveable Camera Video Experiment - Subject 2m away.

Figure 3 shows an example of a subject at a distance of 50 / 60cm from the camera. In this case, the calculated beat value, again considering the average over 30 seconds of video, is 66 bpm, while the real value estimated by the apple watch is 61 bpm.

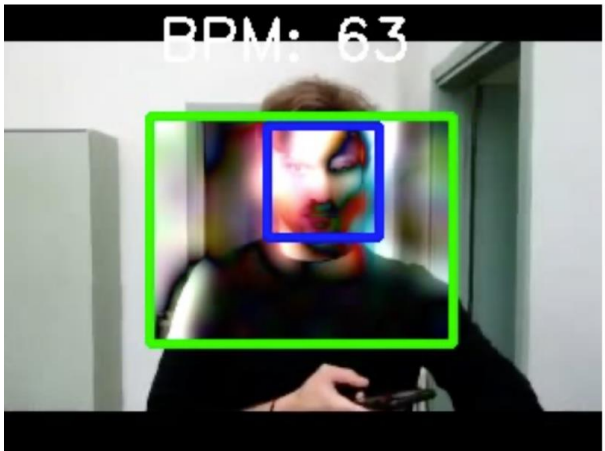

**Fig. 3.** Helmet Video Experiment - Subject 50/60 cm away.

## VI. FSA AND FER ALGORITHMS IMPLEMENTATION

FER and FSA algorithms have quite different implementation logics. While the first requires a single frame extracted from the video to establish the detected emotion, in the FSA algorithm it is necessary, as already highlighted in the previous paragraphs, for a first initial calibration phase, lasting 5 seconds, in which a frame buffer will be populated before feedback can be provided. In addition, enough frames per second must be provided for computation, approximately 15-20.

This substantial difference translates into a longer wait before having feedback on the BPM, as well as unreliability on the returned value when the frames supplied to the algorithm are not enough. The FSA is therefore very dependent on the number of frames passed to the algorithm itself. The greater the number of requests to be performed in a given second, the greater the speed with which the calibration phase will end to provide results as close as possible to the real ones.

Having analyzed the typical work contexts in which the Avatar user will find himself, it has been noticed how with a network that has a download speed of about 6 MBps per second it is possible to perform about 20 requests per minute (1 request every 2-3 seconds), thus making it impossible to have a quick calibration phase and satisfactory results with what has already been implemented. In addition, there are also some limits of Unity (version 2018) that prevent the system from making too many requests without falling into an overflow error and crashing the application itself.

A first possible solution, which could guarantee a turning point to the above-mentioned problem, would be to previously resize (before forwarding) the frame received from the camera and therefore provide for the forwarding of a pool of frames and no more than a single frame, allowing thus, a faster calibration phase and more accurate results.

In this sense, it would be useful from the point of view of the Unity Player, to acquire a defined number of frames (30 frames should be sufficient), and to attempt to make a call to the wrapper API only when the buffer that will contain the frames is full. Once this phase has been completed and a response has been received, it will be possible to continue with the acquisition of an additional block of frames for the next request. This precaution should therefore ensure an improvement in the accuracy of the results obtained by the FSA, introducing a small delay and greater overhead, necessary to accomplish the intended task.

Naturally, such a solution requires considerable refactoring work on the Player Unity side and on the API side, without guaranteeing that the continuation in this direction can solve the problem that arose smoothly.

These changes will also affect the back-end side since it will be necessary to modify the API to accept different frames and then sort them to the reference service, or FER or FSA.

The implementation scheme is shown in Figure 4.

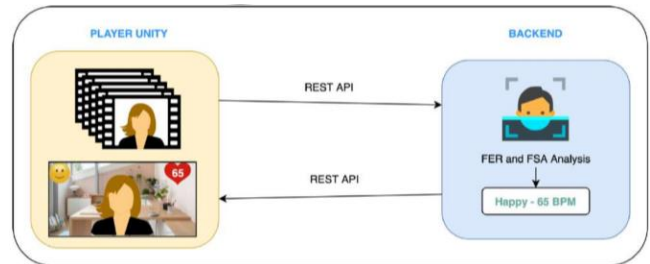

**Fig. 4.** Example of implementation of API Wrapper with sending of n frames per second.

As previously mentioned, the Wrapper API sends the frames once it has collected a good number of them to be able to perform the calculations related to FSA and FER (e.g. 30 frames sent in a single API call, to reduce the load on Unity). At that point, the frames are sent to the backend for processing. When the backend gets the results, it sends them to the frontend for viewing.

Another possible solution to the problem is the implementation of an Apache Kafka architecture [11], which is an architecture that deals with processing large amounts of data in real-time, allowing the creation of scalable systems with high throughput and low latency. Such an architecture is usually used in all those cases where high speed and scalability are essential.

This type of implementation certainly requires a preventive phase to identify any libraries that can be adopted in pre-existing platforms (Player Unity and Back-end) that currently use different languages (C# and Python).

It is also essential to evaluate how and if it is possible to integrate any external libraries not present "in the AssetStore" into the Unity project, ensuring compatibility with all target platforms of the Avatar product (which at the moment are Windows and WebGL). In detail, it is to be considered the remote possibility that any external libraries could give problems to the WebGL build, where the project code is recompiled in JavaScript. It is also essential to evaluate whether these same libraries can be valid choices and easily integrated into what is the current back-end infrastructure, written entirely in Python.

As far as the Player Unity is concerned, the Kafka

architecture should be implemented using the "confluent-kafka-.net" library, available directly on Nuget and therefore importable within the Unity 3D project even if it is not taken directly from the AssetStore. Of course, as previously said, it will still be necessary to evaluate any incompatibilities of a library with the WebGl version of the Avatar project. Confluent-kafka-dotnet is a library, licensed under the Apache license and published via a repository on GitHub, which is maintained and well documented with clear examples explaining how to use it.

It provides both the possibility of implementing producers who are waiting for feedback and "Producers" that (as in our case) can forward streams of requests. It also allows you to implement Consumers, to be able to manage any entities that may receive messages. It is essential to evaluate how to implement external libraries on a Unity project with any incompatibilities in the platforms and it is certainly important to assess whether the library in question can perform the required task in conjunction with those adopted in the back-end.

As for the back-end, it is possible to implement the functionality of Kafka via the Kafka-Python library.

Kafka-Python is a client for the Apache distributed streaming system Kafka, published via a repository on GitHub, is currently kept up-to-date and is distributed under the Apache license. It provides the APIs for the two main functionalities of Kafka, which are the Producer and the Consumer, which work exactly like those present in the official library in java.

Another possibility would be to use the Faust bookcase. Faust is a streaming data processing library, which brings the functionality of Kafka Streams to Python. This was built and is used by Robinhood (a finance company) to build high-performance, real-time distributed systems capable of handling billions of interactions. Being written in python it is easily compatible with other libraries such as NumPy, PyTorch, Pandas, Django, Flask, and SQLAlchemy. Faust requires Python 3.6 or later for the new async / await functions, which is the implementation of asynchronous threads.

Figure 5 shows a graphic description of how this second solution should be implemented. Kafka, in this way, is present both on the Unity side and on the Backend Python side.

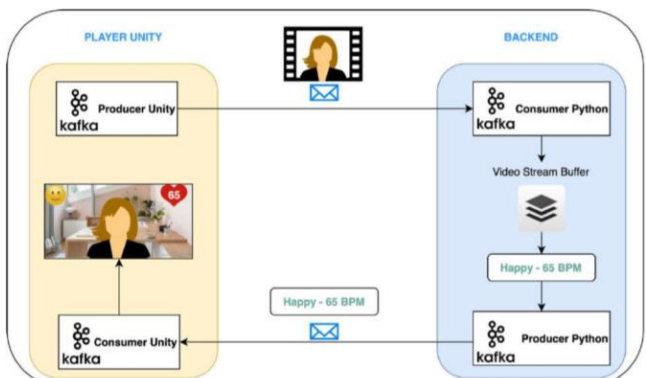

**Fig. 5.** Example of implementation of with Kafka on Unity and Backend side.

A third solution, which starts from the idea expressed in the second solution, is based on creating a similar structure based on Kafka but present only on the server-side and not on the Unity player. This solution has several advantages. First of all, it does not expose the Kafka orchestrator on the web but confines it only to the server and the server's architecture. Secondly, it allows keeping the structure based on RESTful API unaltered on the unity side so as not to compromise, possibly, the WebGL version. The basis would always be the Kafka-Python python library. Figure 7 has represented the architectural idea described above.

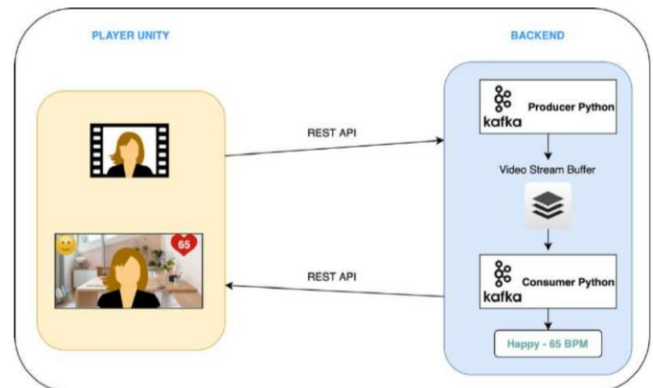

**Fig. 6.** Example of implementation with Kafka on the Backend side.

In this third example, therefore, on the Unity side, the current architecture is left in which, via the REST API, a frame is sent to the backend. At this point, the backend will create a Kafka architecture with producer and consumer in which the first takes the individual frames to create a video stream buffer and sends them to the consumer who must process the data to obtain the output (heartbeat and emotion). The consumer will also take care of providing the data thus obtained directly to the Unity player when these are requested via the REST API.

The integration of these last two functionalities (recognition of facial expressions and estimation of heart rate) has also required a phase of code adjustment to improve performances and to optimize the method of transmission of webcam images from the client (Browser) to the Avatar.xyz server. The second optimization has been necessary above all to limit the traffic of the net and to allow therefore the computation of the level of the beat with reduced latencies.

It was necessary to condense the two applications into a single Unity module that captures the webcam images and that, in parallel, while the recognition of facial expressions is done on the client, sends the images to the server at regular intervals for remote processing.

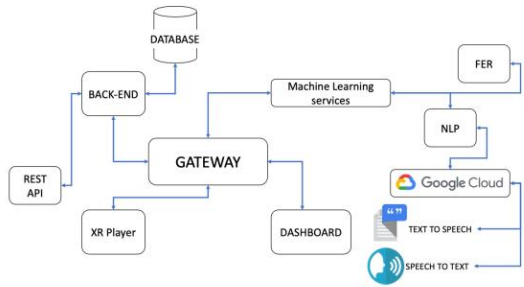

**Fig. 6.** The system architecture.

The description performed here lets us understand how, even starting from methods already approved in laboratory tests, code rewriting steps were necessary to make the systems usable according to the constraints (computational limits, image quality prerequisites, latency times) that emerged during software testing

## VII. CONCLUSION

In this work was performed a solution that provides the learner's training path based on facial expression recognition and facial stress analysis algorithms for the AVATAR platform. By now many application problems of neural networks are solved and available for use, but the actual applicability of these solutions in real contexts requires copious development efforts. During development, facial expression recognition and facial stress analysis algorithms were defined, tested and performance measured. The result of the FSA algorithm work was compared with data from an apple watch and oximeter. An error between the proposed algorithms and the apple watch measurement result is 3.1 beats per minute and the oximeter measurement result is 1 beat. The measurement results are acceptable for the correct operation of the system. Three approaches to the implementation of algorithms in the current platform are also proposed, and their advantages and disadvantages are given. In general, all the presented solutions can be valid for addressing the problem under consideration. It is more feasible to take the third solution that allows not to overturn the architecture that, to date, has been developed. In this way, development times should also decrease and it will not be necessary to find other solutions for the compatibility part with WebGL since communication always takes place via Rest API.